\documentclass[journal,oneside]{IEEEtran}
\usepackage{mathtools}

\usepackage{graphicx}
\usepackage{lettrine}
\usepackage{booktabs}
\usepackage{amsmath}
\usepackage{enumerate}
\usepackage{bm}
\usepackage{graphicx}
\usepackage{epstopdf}
\usepackage{amsmath}
\usepackage{array}
\usepackage{amssymb}
\usepackage{placeins}
\usepackage{cite}
\usepackage{stfloats}

\usepackage{xcolor}
\usepackage{soul}

\newcolumntype{P}[1]{>{\raggedright\arraybackslash}p{#1}}
\newcommand{\tabincell}[2]{\begin{tabular}{@{}#1@{}}#2\end{tabular}}

\ifCLASSINFOpdf
\else
\fi
  \usepackage[caption=false,font=footnotesize]{subfig}
\begin{document}
\title{Short-term Load Forecasting with \\ Deep Residual Networks}

\author{
Kunjin Chen, Kunlong Chen, Qin Wang, Ziyu He, Jun Hu, \emph{Member, IEEE}, and Jinliang He, \emph{Fellow, IEEE} 

\thanks{
K. J. Chen, J. Hu, and J. L. He are with the State Key Lab of Power Systems, Department of Electrical Engineering, Tsinghua University, Beijing 100084, P. R. of China. 

K. L. Chen is with the Department of Electrical Engineering, Beijing Jiaotong University, Beijing 100044, P. R. of China.

Q. Wang is with the Department of Information Technology and Electrical Engineering, ETH Z\"urich, 8092 Z\"urich, Switzerland.

Z. He is with the Department of Industrial and Systems Engineering, University of Southern California, Los Angeles, CA 90007, USA. 

(Corresponding author email: hejl@tsinghua.edu.cn).
} 
}


%Telephone: (800) 555--1212, Fax: (888) 555--1212}

\maketitle
\begin{abstract}
We present in this paper a model for forecasting short-term electric load based on deep residual networks. The proposed model is able to integrate domain knowledge and researchers' understanding of the task by virtue of different neural network building blocks. Specifically, a modified deep residual network is formulated to improve the forecast results. Further, a two-stage ensemble strategy is used to enhance the generalization capability of the proposed model. We also apply the proposed model to probabilistic load forecasting using Monte Carlo dropout. Three public datasets are used to prove the effectiveness of the proposed model. Multiple test cases and comparison with existing models show that the proposed model is able to provide accurate load forecasting results and has high generalization capability.
\end{abstract}

\smallskip
\begin{IEEEkeywords}
Short-term load forecasting, deep learning, deep residual network, probabilistic load forecasting.
\end{IEEEkeywords}


% \fi
\IEEEpeerreviewmaketitle

\section{Introduction}

\IEEEPARstart{T}{he forecasting} of power demand is of crucial importance for the development of modern power systems. The stable and efficient management, scheduling and dispatch in power systems rely heavily on precise forecasting of future loads on various time horizons. In particular, short-term load forecasting (STLF) focuses on the forecasting of loads from several minutes up to one week into the future \cite{ceperic2013strategy}. A reliable STLF helps utilities and energy providers deal with the challenges posed by the higher penetration of renewable energies and the development of electricity markets with increasingly complex pricing strategies in future smart grids. 

Various STLF methods have been proposed by researchers over the years. Some of the models used for STLF include linear or nonparametric regression \cite{song2005short, charytoniuk1998nonparametric}, support vector regression (SVR) \cite{elattar2010electric, ceperic2013strategy}, autoregressive models \cite{taylor2003short}, fuzzy-logic approach \cite{rejc2011short}, etc. Reviews and evaluations of existing methods can be found in \cite{feinberg2005load, taylor2006a, hahn2009electric, wang2018review}. Building STLF systems with artificial neural networks (ANN) has long been one of the main-stream solutions to this task. As early as 2001, a review paper by Hippert \emph{et al}. surveyed and examined a collection of papers that had been published between 1991 and 1999, and arrived at the conclusions that most of the proposed models were over-parameterized and the results they had to offer were not convincing enough \cite{hippert2001neural}. In addition to the fact that the size of neural networks would grow rapidly with the increase in the numbers of input variables, hidden nodes or hidden layers, other criticisms mainly focus on the ``overfitting'' issue of neural networks \cite{ceperic2013strategy}. Nevertheless, different types and variants of neural networks have been proposed and applied to STLF, such as radial basis function (RBF) neural networks \cite{cecati2015novel}, wavelet neural networks \cite{chen2010short, zhao2009short}, extreme learning machines (ELM) \cite{zhang2013short}, to name a few. 

Recent developments in neural networks, especially deep neural networks, have had great impacts in the fields including computer vision, natural language processing, and speech recognition \cite{goodfellow2016deep}. Instead of sticking with fixed shallow structures of neural networks with hand-designed features as inputs, researchers are now able to integrate their understandings of different tasks into the network structures. Different building blocks including convolutional neural networks (CNN) \cite{krizhevsky2012imagenet}, and long short-term memory (LSTM) \cite{hochreiter1997long} have allowed deep neural networks to be highly flexible and effective. Various techniques have also been proposed so that neural networks with many layers can be trained effectively without the vanishing of gradients or severe overfitting. Applying deep neural networks to short-term load forecasting is a relatively new topic. Researchers have been using restricted Boltzmann machines (RBM) and feed-forward neural networks with multiple layers in forecasting of demand side loads and natural gas loads \cite{ryu2016deep, merkel2017deep}. However, these models are increasingly hard to train as the number of layers increases, thus the number of hidden layers are often considerably small (e.g., 2 to 5 layers), which limits the performance of the models.

In this work, we aim at extending existing structures of ANN for STLF by adopting state-of-the-art deep neural network structures and implementation techniques. Instead of stacking multiple hidden layers between the input and the output, we learn from the residual network structure proposed in \cite{he2016deep} and propose a novel end-to-end neural network model capable of forecasting loads of next 24 hours. An ensemble strategy to combine multiple individual networks is also proposed. Further, we extend the model to probabilistic load forecasting by adopting Monte Carlo (MC) dropout (For a comprehensive review of probabilistic electric load forecasting, the reader is referred to \cite{hong2016probabilistic, liu2017probabilistic}). The contributions of this work are three-folds. First, a fully end-to-end model based on deep residual networks for STLF is proposed. The proposed model does not require external feature extraction or feature selection algorithms, and only raw data of loads, temperature and information that is readily available are used as inputs. The results show that the forecasting performance can be greatly enhanced by improving the structure of the neural networks and adopting the ensemble strategy, and that the proposed model has good generalization capability across datasets. To the best of our knowledge, this is the first work that uses deep residual networks for the task of STLF. Second, the building blocks of the proposed model can easily be adapted to existing neural-network-based models to improve forecasting accuracy (e.g., adding residual networks on top of 24-hour forecasts). Third, a new formulation of probabilistic STLF for an ensemble of neural networks is proposed.

The remainder of the paper is organized as follows. In section II, we formulate the proposed model based on deep residual networks. The ensemble strategy, the MC dropout method, as well as the implementation details are also provided. In section III, the results of STLF by the proposed model are presented. We also discuss the performance of the proposed model and compare it with existing methods. Section IV concludes this paper and proposes future works. The source code for the STLF model proposed in this paper is available at \emph{https://github.com/yalickj/load-forecasting-resnet}.

\section{Short-term Load Forecasting Based on Deep Residual Networks}

In this paper, we propose a day-ahead load forecasting model based on deep residual networks. We first formulate the low-level basic structure where the inputs of the model are processed by several fully connected layers to produce preliminary forecasts of 24 hours. The preliminary forecasts are then passed through a deep residual network. After presenting the structure of the deep residual network, some modifications are made to further enhance its learning capability. An ensemble strategy is designed to enhance the generalization capability of the proposed model. The formulation of MC dropout for probabilistic forecasting is also provided.

\subsection{Model Input and the Basic Structure for Load Forecasting of One Hour}

We use the model with the basic structure to give preliminary forecasts of the 24 hours of the next day. Specifically, the inputs used to forecast the load for the $h$th hour of the next day, $L_h$, are listed in Table I. The values for loads and temperatures are normalized by dividing the maximum value of the training dataset. The selected inputs allow us to capture both short-term closeness and long-term trends in the load and temperature time series \cite{zhang2016dnn}. More specifically, we expect that $L^{month}_h$, $L^{week}_h$, $T^{month}_h$ and $T^{week}_h$ can help the model identify long-term trends in the time series (the days of the same day-of-week index as the next day are selected as they are more likely to have similar load characteristics \cite{chen2010short}), while $L^{day}_h$ and $T^{day}_h$ are able to provide short-term closeness and characteristics. The input $L^{hour}_h$ feeds the loads of the most recent 24 hours to the model. Forecast loads are used to replace the values in $L^{hour}_h$ that are not available at the time of forecasting, which also helps associate the forecasts of the whole day. Note that the sizes of the above-mentioned inputs can be adjusted flexibly. In addition, one-hot codes for season\footnote{In this paper, the ranges for Spring, Summer, Autumn, and Winter are March 8th to June 7th, June 8th to September 7th, September 8th to December 7th, December 8th to March 7th, respectively.}, weekday/weekend distinction, and holiday/non-holiday\footnote{In this paper, we consider three major public holidays, namely, Christmas Eve, Thanksgiving Day, and Independence Day as the activities involved in these holidays have great impacts on the loads. The rest of the holidays are considered as non-holidays for simplicity.} distinction are added to help the model capture the periodic and unordinary temporal characteristics of the load time series.

\begin{table}[b]
\renewcommand\arraystretch{1.2}
\centering  % 表居中
\captionsetup{justification=centering}
\caption{Inputs for the Load Forecast of the $h$th Hour of the Next Day} 
\begin{tabular}{l l l}
\toprule[1.5pt]
Input & Size & Description of the Inputs\\
\midrule[0.75pt]
$L^{month}_h$  &  6  &  \tabincell{l}{Loads of the $h$th hour of the days that are 4, 8, 12,\\ 16, 20, and 24 weeks prior to the next day} \\
$L^{week}_h$  &  4  &  \tabincell{l}{Loads of the $h$th hour of the days that are 1, 2, 3,\\ and 4 weeks prior to the next day}          \\
$L^{day}_h$  &  7  &  \tabincell{l}{Loads of the $h$th hour of every day of the week \\prior to the next day}           \\
$L^{hour}_h$  &  24  &  \tabincell{l}{Loads of the most recent 24 hours prior to the $h$th \\hour of the next day}     \\
$T^{month}_h$  &  6  &  \tabincell{l}{Temperature values of the same hours as $L^{month}_h$}           \\
$T^{week}_h$  &  4  &  \tabincell{l}{Temperature values of the same hours as $L^{week}_h$}           \\
$T^{day}_h$  &  7  &  \tabincell{l}{Temperature values of the same hours as $L^{day}_h$}           \\
$T_h$  &  1  &  \tabincell{l}{The actual temperature of the $h$th hour of the next \\ day}            \\
$S$ & 4 & \tabincell{l}{One-hot code for season} \\
$W$ & 2 & \tabincell{l}{One-hot code for weekday/weekend distinction} \\
$H$ & 2 & \tabincell{l}{One-hot code for holiday/non-holiday distinction} \\
\bottomrule[1.5pt]
\end{tabular}
\end{table}

\begin{figure}[tb]
\centering
\includegraphics[width=8.7cm]{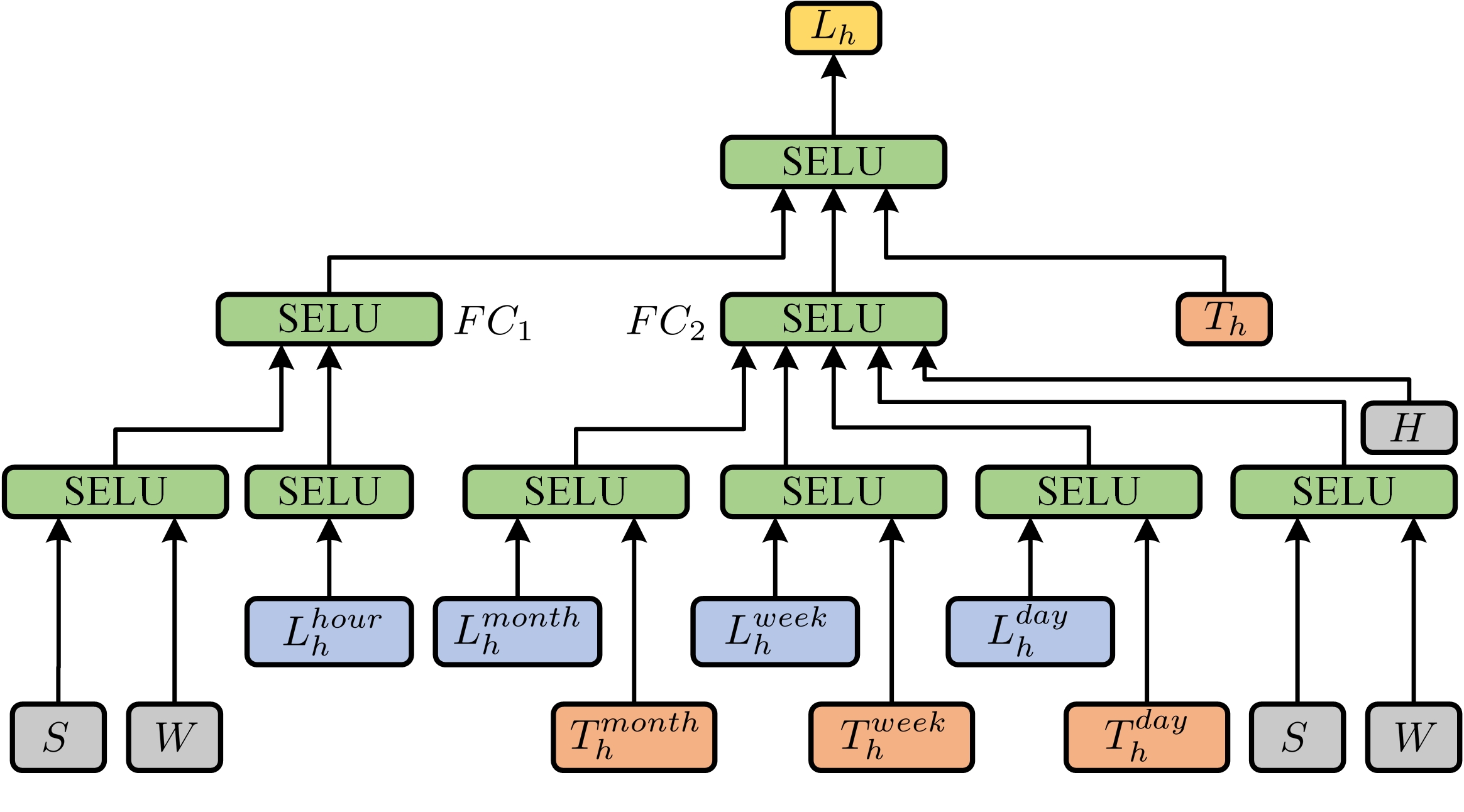}
\caption{The structure of the neural network model for load forecasting of one hour.}
\label{one_hour}
\end{figure}

The structure of the neural network model for load forecasting of one hour is illustrated in Fig. \ref{one_hour}. For $L^{month}_h$, $L^{week}_h$, $L^{day}_h$, $T^{month}_h$, $T^{week}_h$, and $T^{day}_h$, we first concatenate the pairs $[L^{month}_h,T^{month}_h]$, $[L^{week}_h,T^{week}_h]$, and $[L^{day}_h,T^{day}_h]$, and connect them with three separate fully-connected layers. The three fully-connected layers are then concatenated and connected with another fully-connected layer denoted as $FC_2$. For $L^{hour}_h$, we forward pass it through two fully-connected layers, the second layer of which is denoted as $FC_1$.
$S$ and $W$ are concatenated to produce two fully-connected layers, one used as part of the input of $FC_1$, the other used as part of the input of $FC_2$. $H$ is also connected to $FC_2$. In order to produce the output $L_h$, we concatenate $FC_1$, $FC_2$, and $T_h$, and connect them with a fully-connected layer. This layer is then connected to $L_h$ with another fully connected layer. All fully-connected layers but the output layer use scaled exponential linear units (SELU) as the activation function.

The adoption of the ReLU has greatly improved the performance of deep neural networks \cite{dahl2013improving}. Specifically, ReLU has the form  

\begin{equation}
	\text{ReLU}(y_i) = \max(0, y_i)
\end{equation}
where $y_i$ is the linear activation of the $i$-th node of a layer. A problem with ReLU is that if a unit can not be activated by any input in the dataset, the gradient-based optimization algorithm is unable to update the weights of the unit, so that the unit will never be activated again. In addition, the network will become very hard to train if a large proportion of the hidden units produce constant 0 gradients \cite{maas2013rectifier}. This problem can be solved by adding a slope to the negative half axis of ReLU. With a simple modification to the formulation of ReLU on the negative half axis, we get PReLU \cite{he2015delving}. The activations of a layer with PReLU as the activation function is obtained by
\begin{equation}
	\text{PReLU}(y_i) = 
	\begin{cases}
		y_i &\mbox{if $y_i>0$}\\
		\beta_i y_i &\mbox{if $y_i \leq 0$}
   \end{cases}
\end{equation}
where $\beta_i$ is the coefficient controlling the slope of $\beta_iy_i$ when $y_i \leq 0$. A further modification to ReLU that induces self-normalizing properties is provided in \cite{klambauer2017self}, where the activation function of SELU is given by 

\begin{equation}
	\text{SELU}(y_i) = \lambda
	\begin{cases}
		y_i &\mbox{if $y_i>0$}\\
		\alpha e^{y_i} - \alpha &\mbox{if $y_i \leq 0$}
   \end{cases}
\end{equation}
where $\lambda$ and $\alpha$ are two tunable parameters. It is shown in \cite{klambauer2017self} that if we have $\lambda \approx 1.0577$ and $\alpha \approx 1.6733$, the outputs of the layers in a fully-connected neural network would approach the standard normal distribution when the inputs follow the standard normal distribution. This helps the networks to prevent the problems of vanishing and exploding gradients.

As previously mentioned, in order to associate the forecasts of the 24 hours of the next day, the corresponding values within $L^{hour}_h$ are replaced by $\left\lbrace L_1, \cdots, L_{h-1} \right\rbrace$ for $h>1$. Instead of simply copying the values, we maintain the neural network connections underneath them. Thus, the gradients of subsequent hours can be propagated backward through time. This would help the model adjust the forecast value of each hour given the inputs and forecast values of the rest of the hours.

We then concatenate $\left\lbrace L_1,\cdots,L_{24} \right\rbrace$ as $L$, which directly becomes the output of the model with the basic structure. Next, we proceed to formulate the deep residual network and add it on top of $L$. The output of the deep residual network is denoted as $\hat{y}$ and has the same size of $L$.

\subsection{The Deep Residual Network Structure for Day-ahead Load Forecasting}

\begin{figure}[tb]
\centering
\includegraphics[width=3.2cm]{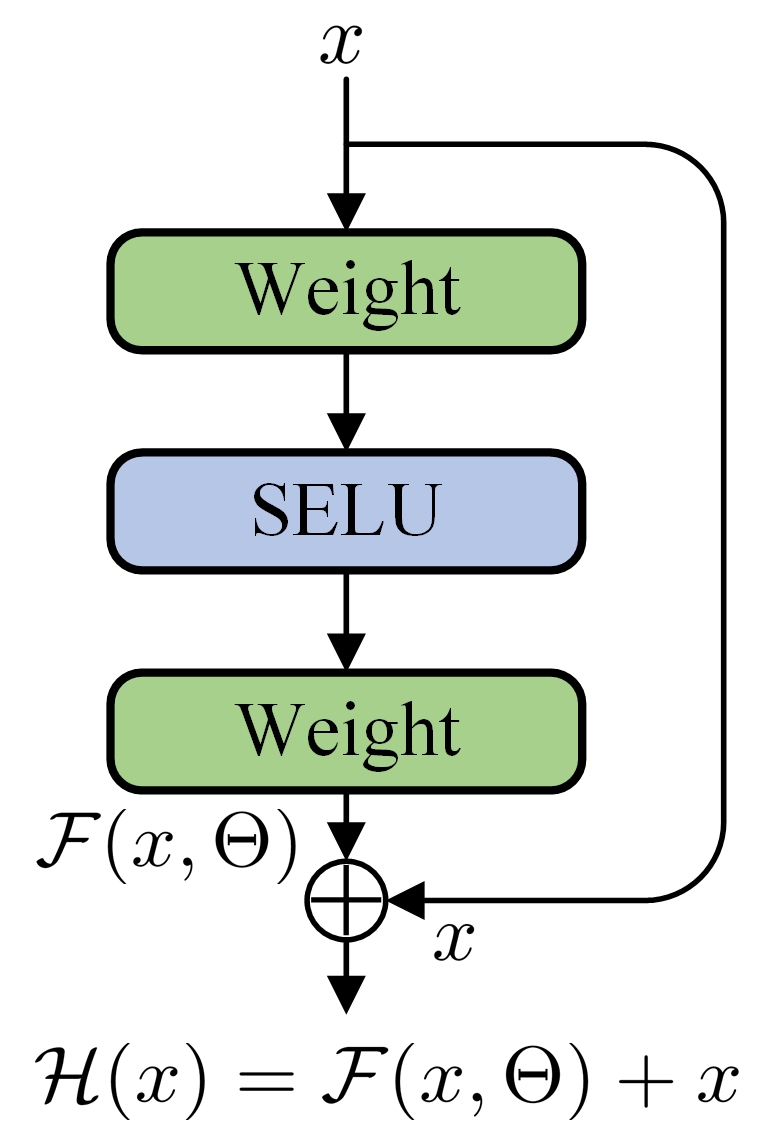}
\caption{The building block of the deep residual network. SELU is used as the activation function between two linear layers.}
\label{res_block}
\end{figure}

\begin{figure}[tb]
\centering
\includegraphics[width=3cm]{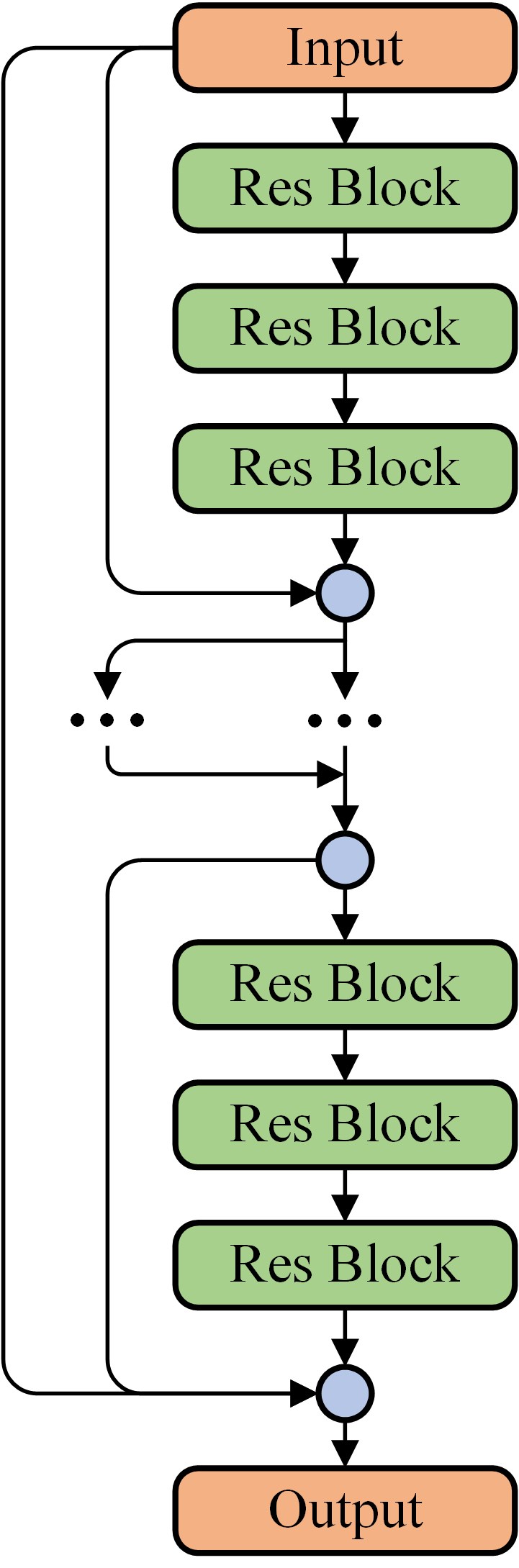}
\caption{An illustration of the deep residual network (ResNet) structure. More shortcut connections are made in addition to the ones within the blocks. In this figure, every three residual blocks has one shortcut connection and another shortcut connection is made from the input to the output. Each round node averages all of its inputs.}
\label{resnet}
\end{figure}

In \cite{he2016deep}, an innovative way of constructing deep neural networks for image recognition is proposed. In this paper, the residual block in Fig. \ref{res_block} is used to build the deep neural network structure. In the residual block, instead of learning a mapping from $x$ to $\mathcal{H}(x)$, a mapping from $x$ to $\mathcal{F}(x,\Theta)$ is learned, where $\Theta$ is a set of weights (and biases) associated with the residual block. Thus, the overall representation of the residual block becomes

\begin{equation}
\mathcal{H}(x) = \mathcal{F}(x,\Theta) + x
\end{equation}
A deep residual network can be easily constructed by stacking a number of residual blocks. We illustrate in Fig. \ref{resnet} the structure of the deep residual network (ResNet) used for the proposed model. More specifically, if $K$ residual blocks are stacked, the forward propagation of such a structure can be represented by

\begin{equation}
x_K = x_0 + \sum_{i=1}^{K}{\mathcal{F}(x_{i-1}, \Theta_{i-1})}
\end{equation}
where $x_0$ is the input of the residual network, $x_K$ the output of the residual network, and $\Theta_i=\left\lbrace \theta_{i,l}|_{1\leq l\leq L} \right\rbrace$ the set of weights associated with the $i$th residual block, $L$ being the number of layers within the block. The back propagation of the overall loss of the neural network to $x_0$ can then be calculated as

\begin{equation}
\frac{\partial \mathcal{L}}{\partial x_0} = \frac{\partial \mathcal{L}}{\partial x_K}(1+\frac{\partial }{\partial x_0}\sum_{i=1}^{K}{\mathcal{F}(x_{i-1}, \Theta_{i-1})})
\end{equation}
where $\mathcal{L}$ is the overall loss of the neural network. The ``1'' in the equation indicates that the gradients at the output of the network can be directly back-propagated to the input of the network, so that the vanishing of gradients (which is often observed when the gradients at the output have to go through many layers before reaching the input) in the network is much less likely to occur \cite{he2016identity}. As a matter of fact, this equation can also be applied to any pair $(x_i, x_j)$ ($0 \leq i<j \leq K$), where $x_i$ and $x_j$ are the output of the $i$th residual block (or the input of the network when $i=0$), and the $j$th residual block, respectively.

In addition to the stacked residual blocks, extra shortcut connections can be added into the deep residual network, as is introduced in \cite{zhang2017residual}. Concretely, two levels of extra shortcut connections are added to the network. The lower level shortcut connection bypasses several adjacent residual blocks, while the higher level shortcut connection is made between the input and output. If more than one shortcut connection reaches a residual block or the output of the network, the values from the connections are averaged. Note that after adding the extra shortcut connections, the formulations of the forward-propagation of responses and the back-propagation of gradients are slightly different, but the characteristics of the network that we care about remain unchanged.  

\begin{figure}[tb]
\centering
\includegraphics[width=7cm]{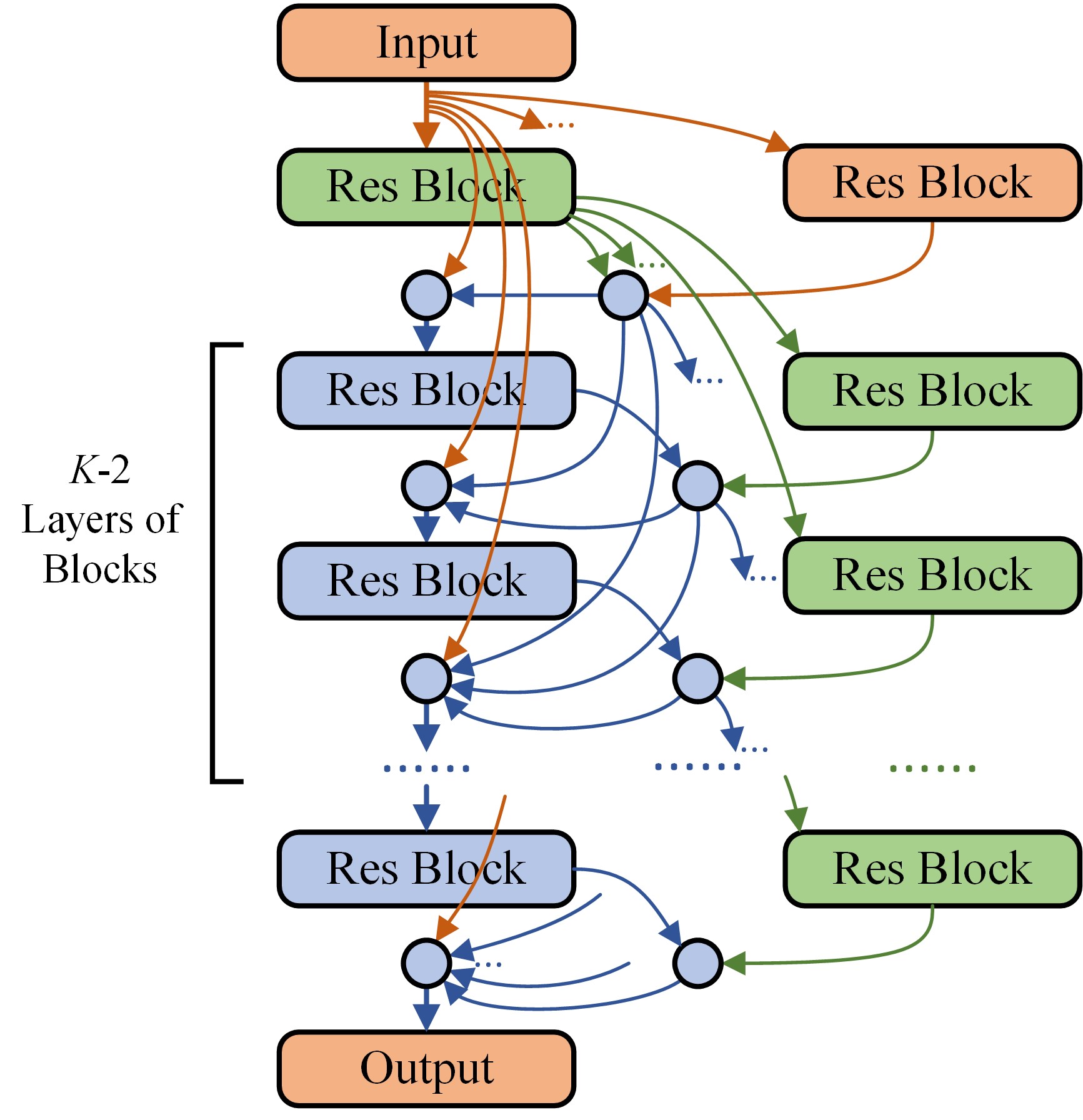}
\caption{An illustration of the modified deep residual network (ResNetPlus) structure. The blue dots in the figure average their inputs, and the outputs are connected to subsequent residual blocks.}
\label{resnetplus}
\end{figure}

We can further improve the learning ability of ResNet by modifying its structure. Inspired by the convolutional network structures proposed in \cite{huang2017densely, zhao2016connection}, we propose the modified deep residual network (ResNetPlus), whose structure is shown in Fig. \ref{resnetplus}. First, we add a series of side residual blocks to the model (the residual blocks on the right). Unlike the implementation in \cite{zhao2016connection}, the input of the side residual blocks is the output of the first residual block on the main path (except for the first side residual block, whose input is the input of the network). The output of each main residual block is averaged with the output of the side residual block in the same layer (indicated by the blue dots on the right). Similar to the densely connected network in \cite{huang2017densely}, the outputs of those blue dots are connected to all main residual blocks in subsequent layers. Starting from the second layer, the input of each main residual block is obtained by averaging all connections from the blue dots on the right together with the connection from the input of the network (indicated by the blue dots on the main path). It is expected that the additional side residual blocks and the dense shortcut connections can improve the representation capability and the efficiency of error back-propagation of the network. Later in this paper, we will compare the performance of the basic structure, the basic structure connected with ResNet, and the basic structure connected with ResNetPlus.

\subsection{The Ensemble Strategy of Multiple Models}

\begin{figure}[tb]
\centering
\includegraphics[width=8.5cm]{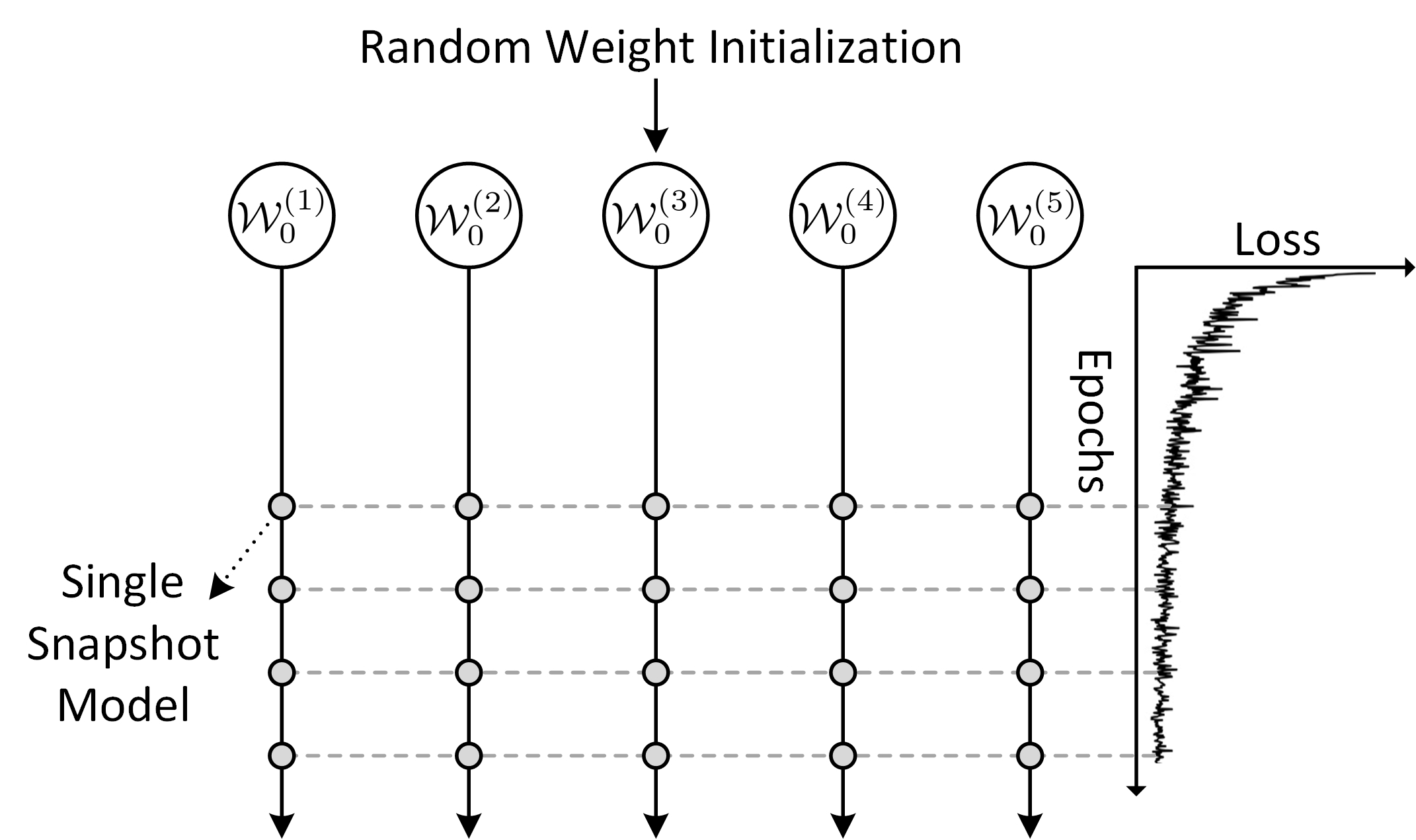}
\caption{A demonstration of the ensemble strategy used in this paper. The snapshot models are taken where the slope of validation loss is considerably small.}
\label{ensemble}
\end{figure}

It is widely acknowledged in the field of machine learning that an ensemble of multiple models has higher generalization capability \cite{goodfellow2016deep} than individual models. In \cite{de2011short}, analysis of neural network ensembles for STLF of office buildings is provided by the authors. Results show that an ensemble of neural networks reduces the variance of performances. A demonstration of the ensemble strategy used in this paper is shown in Fig. \ref{ensemble}. More specifically, the ensemble strategy consists of two stages.

The first stage of the strategy takes several snapshots during the training of a single model. In \cite{huang2017snapshot}, the authors show that setting cyclic learning rate schedules for stochastic gradient descent (SGD) optimizer greatly improves the performance of existing deep neural network models. In this paper, as we use Adam (abbreviated from adaptive moment estimation \cite{kingma2017adam}) as the optimizer, the learning rates for each iteration are decided adaptively. Thus, no learning rate schedules are set by ourselves. This scheme is similar to the NoCycle snapshot ensemble method discussed in \cite{huang2017snapshot}, that is, we take several snapshots of the same model during its training process (e.g., the 4 snapshots along the training process of the model with initial parameters $W_0^{(1)}$). As is indicated in Fig. \ref{ensemble}, the snapshots are taken after an appropriate number of epochs, so that the loss of each snapshot is of similar level.

We can further ensemble a number of models that are trained independently. This is done by simply re-initializing the parameters of the model (e.g., $\mathcal{W}_0^{(1)}$ to $\mathcal{W}_0^{(5)}$ are 5 sets of initial parameters sampled from the same distribution used for initializing the model), which is one of the standard practices of obtaining good ensemble models \cite{webb2003statistical}. The numbers of snapshots and re-trained models are hyper-parameters, which means they can be tuned using the validation dataset. After we obtain the all the snapshot models, we average the outputs of the models and produce the final forecast.

\subsection{Probabilistic Forecasting Based on Monte Carlo Dropout}
If we look at the deep residual network (either ResNet or ResNetPlus) as an ensemble of relatively shallow networks, the increased width and number of connections in the network can provide more shallow networks to form the ensemble model \cite{zhao2016connection}. It is expected that the relatively shallow networks themselves can partially capture the nature of the load forecasting task, and multiple shallow networks with the same input can give varied outputs. This indicates that the proposed model have the potential to be used for probabilistic load forecasting.

Probabilistic forecasting of time series can be fulfilled by capturing the uncertainty within the models \cite{zhu2017deep}. From a Bayesian probability theory point of view, the predictive probability of a Bayesian neural network can be obtained with

\begin{equation}
p(y^*|x^*) = \int_{\mathcal{W}}p(y^*|f^{\mathcal{W}}(x^*))p(\mathcal{W}|X,Y)\,\mathrm{d}\mathcal{W}
\end{equation}
where $X$ and $Y$ are the observations we use to train $f^{\mathcal{W}}(\cdot)$, a neural network with parameters $\mathcal{W}$. The intractable posterior distribution $p(\mathcal{W}|X,Y)$ is often approximated by various inference methods \cite{zhu2017deep}. In this paper, we use MC dropout \cite{gal2016dropout} to obtain the probabilistic forecasting uncertainty, which is easy and computationally efficient to implement. Specifically, dropout refers to the technique of randomly dropping out hidden units in a neural network during the training of the network \cite{srivastava2014dropout}, and a parameter $p$ is used to control the probability that any hidden neuron is dropped out. If we apply dropout stochastically for $M$ times at test time and collect the outputs of the network, we can approximate the first term of the forecasting uncertainty, which is
\begin{equation}
\begin{aligned}
\mathrm{Var}(y^*|x^*) & =\mathrm{Var}\left[\mathbb{E}(y^*|\mathcal{W},x^*)\right]+\mathbb{E}\left[\mathrm{Var}(y^*|\mathcal{W},x^*)\right] \\
& = \mathrm{Var}(f^{\mathcal{W}}(x^*))+ \sigma^2 \\
& \approx \frac{1}{M} \sum^M_{m=1}(\hat{y}^*_{(m)}-\bar{\hat{y}}^*)^2 + \sigma^2
\end{aligned}
\end{equation}
where $\hat{y}^*_{(m)}$ is the $m$th output we obtain, $\bar{\hat{y}}^*$ is the mean of all $M$ outputs, and $\mathbb{E}$ denotes the expectation operator. The second term, $\sigma^2$, measures the \emph{inherent noise} for the data generating process. According to \cite{zhu2017deep}, $\sigma^2$ can be estimated using an independent validation dataset. We denote the validation dataset with $X'=\left\lbrace x'_1, \cdots, x'_V \right\rbrace$, $Y'=\left\lbrace y'_1, \cdots, y'_V \right\rbrace$, and estimate $\sigma^2$ by
\begin{equation}
\sigma^2 = \frac{\beta}{V}\sum^V_{v=1}{(y'_v - f^{\hat{\mathcal{W}}}(x'_v))^2}
\end{equation}
where $f^{\hat{\mathcal{W}}}(\cdot)$ is the model trained on the training dataset and $\beta$ is a parameter to be estimated also using the validation dataset. 

We need to extend the above estimation procedure to an ensemble of models. Concretely, for an ensemble of $K$ neural network models of the same structure, we estimate the first term of (8) with a single model of the same structure trained with dropout. The parameter $\beta$ in (9) is also estimated by the model. More specifically, we find the $\beta$ that provides the best 90$\%$ and 95$\%$ interval forecasts on the validation dataset. $\sigma^2$ is estimated by replacing $f^{\hat{\mathcal{W}}}(\cdot)$ in (9) by the ensemble model, $f^*(\cdot)$. Note that the estimation of $\sigma^2$ is specific to each hour of the day. 

After obtaining the forecasting uncertainty for each forecast, we can calculate the $\alpha$-level interval with the point forecast, $f^*(x^*)$, and its corresponding quantiles to obtain probabilistic forecasting results.

\subsection{Model Design and Implementation Details}
The proposed model consists of the neural network structure for load forecasting of one hour (referred to as the basic structure), the deep residual network (referred to as ResNet) for improving the forecasts of 24 hours, and the modified deep residual network (referred to as ResNetPlus). The configurations of the models are elaborated as follows.

\subsubsection{The model with the basic structure}
The graphic representation of the model with the basic structure is shown in Fig. \ref{one_hour}. Each fully-connected layer for $[L^{day}_h,T^{day}_h]$, $[L^{week}_h,T^{week}_h]$, $[L^{month}_h,T^{month}_h]$, and $L^{hour}_h$ has 10 hidden nodes, while the fully-connected layers for $[S,W]$ have 5 hidden nodes. $FC_1$, $FC_2$, and the fully-connected layer before $L_h$ have 10 hidden nodes. All but the output layer use SELU as the activation function.

\subsubsection{The deep residual network (ResNet)}
ResNet is added to the neural network with the basic structure. Each residual block has a hidden layer with 20 hidden nodes and SELU as the activation function. The size of the outputs of the blocks is 24, which is the same as that of the inputs. A total of 30 residual blocks are stacked, forming a 60-layer deep residual network. The second level of shortcut connections is made every 5 residual blocks. The shortcut path of the highest level connects the input and the output of the network. 

\subsubsection{The modified deep residual network (ResNetPlus)}
The structure of ResNetPlus follows the structure shown in Fig. \ref{resnetplus}. The hyper-parameters inside the residual blocks are the same as ResNet.

In order to properly train the models, the loss of the model, $\mathcal{L}$, is formulated as the sum of two terms:

\begin{equation}
\mathcal{L} = \mathcal{L}_{E} + \mathcal{L}_{R}
\end{equation}
where $\mathcal{L}_{E}$ measures the error of the forecasts, and $\mathcal{L}_{R}$ is an out-of-range penalty term used to accelerate the training process. Specifically, $\mathcal{L}_{E}$ is defined as

\begin{equation}
\mathcal{L}_{E} = \frac{1}{NH}\sum^N_{i=1}{\sum^H_{h=1}{\frac{\left|\hat{y}_{(i,h)} - y_{(i,h)}\right|}{y_{(i,h)}}}} 
\end{equation}
where $\hat{y}_{(i,h)}$ and $y_{(i,h)}$ are the output of the model and the actual normalized load for the $h$th hour of the $i$th day, respectively, $N$ the number of data samples, and $H$ the number of hourly loads within a day (i.e., $H=24$ in this case). This error measure, widely known as the mean absolute percentage error (MAPE), is also used to evaluate the forecast results of the models. The second term, $\mathcal{L}_{R}$, is calculated as

\begin{equation}
\begin{aligned}
\mathcal{L}_{R} = & \frac{1}{2N}\sum^N_{i=1} \max(0, \max\limits_h\hat{y}_{(i,h)} - \max\limits_hy_{(i,h)})  \\ 
& + \max(0, \min\limits_hy_{(i,h)} - \min\limits_h\hat{y}_{(i,h)})
\end{aligned}
\end{equation}
This term penalizes the model when the forecast daily load curves are out of the range of the actual load curves, thus accelerating the beginning stage of the training process. When a model is able to produce forecasts with relatively high accuracy, this term serves to emphasize the cost for overestimating the peaks and the valleys of the load curves.

All the models are trained using the Adam optimizer with default parameters as suggested in \cite{kingma2017adam}. The models are implemented using \texttt{Keras} 2.0.2 with \texttt{Tensorflow} 1.0.1 as backend in the Python 3.5 environment \cite{chollet2015keras, abadi2016tensorflow}. A laptop with Intel$\textregistered$ Core$^{\rm{TM}}$ i7-5500U CPUs is used to train the models. Training the ResNetPlus model with data of three years for 700 epochs takes approximately 1.5 hours. When 5 individual models are trained, the total training time is less than 8 hours. 

\section{Results and Discussion}

In this section, we use the North-American Utility dataset\footnote{Available at https://class.ee.washington.edu/555/el-sharkawi.} and the ISO-NE dataset\footnote{Available at https://www.iso-ne.com/isoexpress/web/reports/load-and-demand.} to verify the effectiveness of the proposed model. As we use actual temperature as the input, we further modify the temperature values to evaluate the performance of the proposed model. Results of probabilistic forecasting on the North-American Utility dataset and the GEFCom2014 dataset \cite{Hong2016P} are also provided.

\subsection{Performance of the Proposed model on the North-American Utility Dataset}

The first test case uses the North-American Utility dataset. This dataset contains load and temperature data at one-hour resolution for a north-American utility. The dataset covers the time range between January 1st, 1985 and October 12th, 1992. The data of the two-year period prior to October 12th, 1992 is used as the test set, and the data prior to the test set is used for training the model. More specifically, two starting dates, namely, January 1st, 1986, and January 1st, 1988, are used for the training sets. As the latter starting date is used in experiments in the literature, we tune the hyper-parameters using the last 10$\%$ of the training set with this starting date\footnote{For this dataset, 4 snapshots are taken between 1200 to 1350 epochs for 8 individual models. For the basic structure, all layers except the input and the output layers are shared for the 24 hours (sharing weights for 24 hours is only implemented in this test case). The ResNetPlus model has 30 layers on the main path.}. The model trained with the training set containing 2 years of extra data has the same hyper-parameters. 

\begin{figure}[tb]
\setlength{\abovecaptionskip}{0pt}
\centering
\includegraphics[width=7cm]{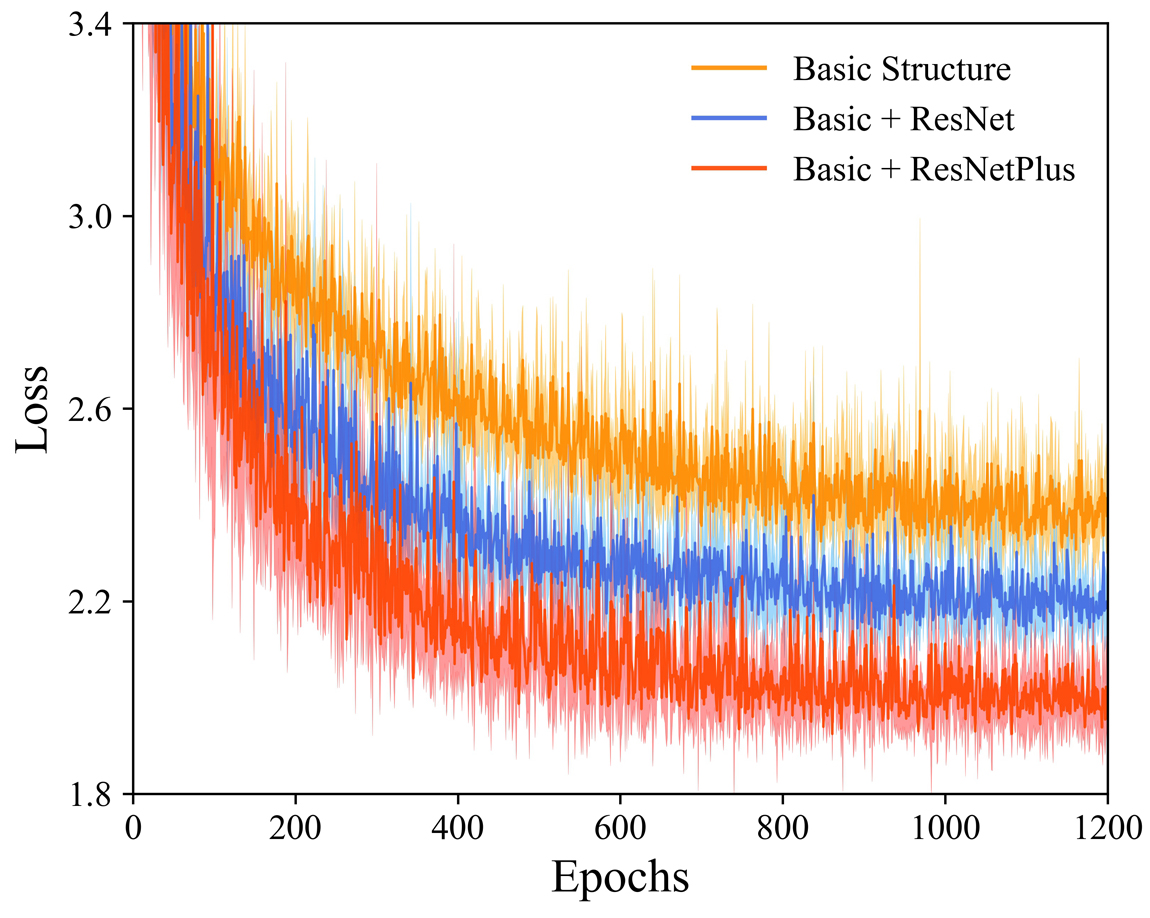}
\caption{Test losses of the neural network with the basic structure (Basic), the model with the deep residual network (Basic + ResNet), and the model with the modified deep residual network (Basic + ResNetPlus). Each model is trained 5 times with shuffled weight initialization. The solid lines are the average losses, and the standard deviation above and below the average losses are indicated by coloured areas.}
\label{test_loss}
\end{figure}

Before reporting the performance of the ensemble model obtained by combining multiple individual models, we first look at the performance of the three models mentioned in section II. The test losses of the three models are shown in Fig. \ref{test_loss} (the models are trained with the training set starting with January 1st, 1988). In order to yield credible results, we train each model 5 times and average the losses to obtain the solid lines in the figure. The coloured areas indicate the range between one standard deviation above and below the average losses. It is observed in the figure that ResNet is able to improve the performance of the model, and further reduction in loss can be achieved when ResNetPlus is implemented. Note that the results to be reported in this paper are all obtained with the ensemble model. For simplicity, the ensemble model with the basic structure connected with ResNetPlus is referred to as ``the ResNetPlus model'' hereinafter.

\begin{table}[!tb]
\renewcommand\arraystretch{1}
\centering  % 表居中
\captionsetup{justification=centering}
\caption{Comparison of the Proposed ResNetPlus Model with Existing Models on the North-American Utility Dataset with Respect to MAPE ($\%$) } 
\begin{tabular}{p{3.5cm} p{1.7cm} p{1.7cm}}
\toprule[1.5pt]

Model & \tabincell{l}{Actual \\temperature} & \tabincell{l}{Modified \\temperature}  \\
 
\midrule[0.75pt]
WT-NN \cite{reis2005feature}  & $2.64$ &  $2.84$           \\
WT-NN \cite{Amjady2009Short}  & $2.04$ &  -           \\
ESN \cite{deihimi2012application}  & $2.37$ &  $2.53$           \\
SSA-SVR \cite{ceperic2013strategy} & $1.99$ &  $2.03$           \\
WT-ELM-MABC \cite{li2015short} & $1.87$ & $1.95$ \\
CLPSO-MA-SVR \cite{hu2014comprehensive} & $1.80$ &  $1.85$           \\
WT-ELM-LM \cite{li2016novel} & $1.67$ &  $1.73$           \\
Proposed model  & $\bf{1.665}$ &  $\bf{1.693}$           \\
Proposed model (2 extra years)  & $\bf{1.557}$ &  $\bf{1.575}$           \\

\bottomrule[1.5pt]
\end{tabular}
\label{north_american}
\end{table}

We compare the results of the proposed ResNetPlus model with existing models proposed in \cite{ceperic2013strategy, reis2005feature, Amjady2009Short, deihimi2012application, li2015short, hu2014comprehensive, li2016novel}, as is shown in Table \ref{north_american}. In order to estimate the performance of the models when forecast temperature is used, we also add a Gaussian noise with mean 0 $^\text{o}$F, and standard deviation 1 $^\text{o}$F to the temperature input  and report the MAPE in this case. It is seen in the table that the proposed model outperforms existing models which highly depend on external feature extraction, feature selection, or hyper-parameter optimization techniques. The proposed model also has a lower increase of MAPE when modified temperature is applied. In addition, the test loss can be further reduced when more data is added to the training set.

\subsection{Performance of the Proposed Model on the ISO-NE Dataset}

The second task of the paper is to examine the generalization capability of the proposed model. To this end, we use the majority of the hyper-parameters of ResNetPlus tuned with the North-American Utility dataset to train load forecasting models for the ISO-NE dataset (The time range of the dataset is between March 2003 and December 2014). Here, the ResNetPlus structure has 10 layers on the main path.

The first test case is to predict the daily loads of the year 2006 in the ISO-NE dataset. For the proposed ResNetPlus model, the training period is from June 2003 to December 2005\footnote{The training dataset is used to determine how the snapshots are taken for the ensemble model for the ISO-NE dataset. For each implementation, 5 individual models are trained, and the snapshots are taken at 600, 650, and 700 epochs.} (we reduce the size of $L^{month}_h$ and $T^{month}_h$ to 3 so that more training samples can be used, and the rest of the hyper-parameters are unchanged). In comparison, the similar day-based wavelet neural network (SIWNN) model in \cite{chen2010short} is trained with data from 2003 to 2005, while the models proposed in \cite{li2016ensemble} and \cite{li2015short} use data from March 2003 to December 2005 (both models use past loads up to 200 hours prior to the hour to be predicted). The results of MAPEs with respect to each month are listed in Table \ref{ISO2006}. The MAPEs for the 12 months in 2006 are not explicitly reported in \cite{li2016ensemble}. It is seen in the table that the proposed ResNetPlus model has the lowest overall MAPE for the year 2006. For some months, however, the WT-ELM-MABC model proposed in \cite{li2015short} produces better results. Nevertheless, as most of the hyper-parameters are not tuned on the ISO-NE dataset, we can conclude that the proposed model has good generalization capability across different datasets.

We further test the generalization capability of the proposed ResNetPlus model on data of the years 2010 and 2011. The same model for the year 2006 is used for this test case, and historical data from 2004 to 2009 is used to train the model. In Table \ref{ISO2010}, we report the performance of the proposed model and compare it with models mentioned in \cite{yu2014incremental, cecati2015novel, li2016ensemble}. Results show that the proposed ResNetPlus model outperforms existing models with respect to the overall MAPE for the two years, and an improvement of 8.9\% is achieved for the year 2011. Note that all the existing models are specifically tuned on the ISO-NE dataset for the period from 2004 to 2009, while the design of the proposed ResNetPlus model is directly implemented without any tuning.

\begin{table}[!tb]
\renewcommand\arraystretch{1}
\centering  % 表居中
\captionsetup{justification=centering}
\caption{MAPEs ($\%$) of the Proposed ResNetPlus Model for the ISO-NE Dataset in 2006 and A Comparison with Existing Models} 
\begin{tabular}{P{0.8cm} P{1.3cm} P{1.3cm} P{1.3cm} P{1.3cm}}
\toprule[1.5pt]

& SIWNN \cite{chen2010short} & WT-ELM-PLSR \cite{li2016ensemble} & WT-ELM-MABC \cite{li2015short} & Proposed model \\
 
\midrule[0.75pt]
Jan  & $1.60$      &  -  &  $\bf{1.52}$  &  $1.619$         \\
Feb  & $1.43$      &  -  &  $\bf{1.28}$  &  $1.308$         \\
Mar  & $1.47$      &  -  &  $1.37$       &  $\bf{1.172}$         \\
Apr  & $1.26$      &  -  &  $\bf{1.05}$  &  $1.340$         \\
May  & $1.61$      &  -  &  $\bf{1.23}$  &  $1.322$         \\
Jun  & $1.79$      &  -  &  $1.54$       &  $\bf{1.411}$         \\
Jul  & $2.70$      &  -  &  $2.07$       &  $\bf{1.962}$         \\
Aug  & $2.62$      &  -  &  $2.06$       &  $\bf{1.549}$         \\
Sep  & $1.48$      &  -  &  $1.41$       &  $\bf{1.401}$         \\
Oct  & $1.38$      &  -  &  $\bf{1.23}$  &  $1.293$        \\
Nov  & $1.39$      &  -  &  $\bf{1.33}$  &  $1.507$         \\
Dec  & $1.75$      &  -  &  $1.65$       &  $\bf{1.465}$         \\
\midrule[0.75pt]
Average & $1.75$ & $ 1.489 $ & $ 1.48 $ & $ \bf{1.447} $ \\
\bottomrule[1.5pt]
\end{tabular}
\label{ISO2006}
\end{table}

\begin{table}[!tb]
\renewcommand\arraystretch{1}
\centering  % 表居中
\captionsetup{justification=centering}
\caption{Comparison of the Proposed ResNetPlus Model with Existing Models on the ISO-NE dataset for 2010 and 2011} 
\begin{tabular}{p{3.5cm} p{1.7cm} p{1.7cm}}
\toprule[1.5pt]

Model & 2010 & 2011 \\
 
\midrule[0.75pt]
RBFN-ErrCorr original \cite{yu2014incremental} & $1.80$ & $2.02$\\
RBFN-ErrCorr modified \cite{cecati2015novel} & $1.75$ & $1.98$\\
WT-ELM-PLSR \cite{li2016ensemble} & $\bf{1.50}$ &  $1.80$           \\
Proposed model  & $\bf{1.50}$ &  $\bf{1.64}$           \\

\bottomrule[1.5pt]
\end{tabular}
\label{ISO2010}
\end{table}

\begin{figure}[!htbp]
\setlength{\abovecaptionskip}{0pt}
\centering
\includegraphics[width=7cm]{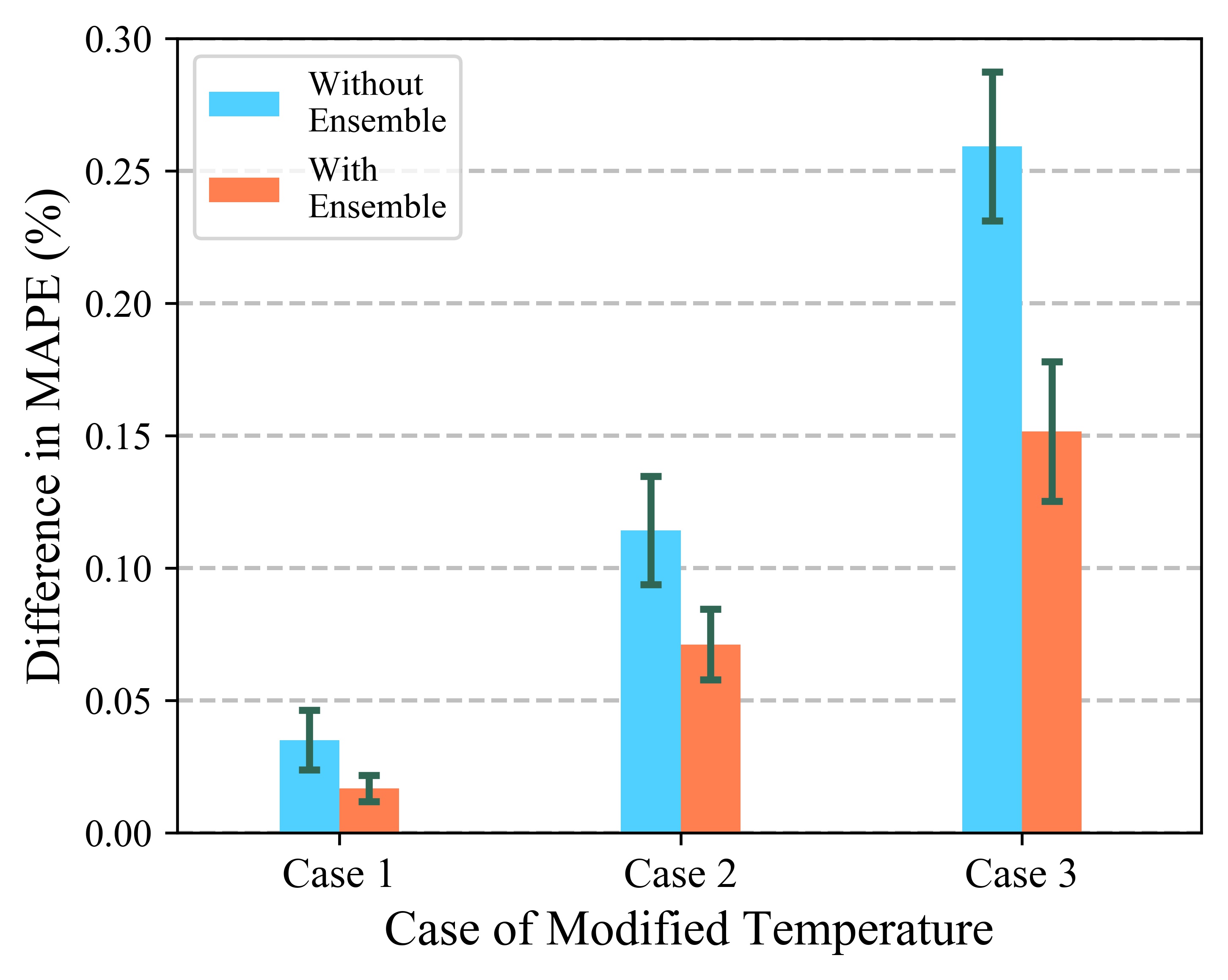}
\caption{The comparison of the proposed model with the ensemble strategy and the proposed model without ensemble when different cases of modified temperature are applied. The model without ensemble is a single ResNetPlus model trained with 700 epochs.}
\label{temp}
\end{figure}

As we use actual temperature values for the input of the proposed model (except for the "modified temperature" case of North-American Utility dataset), the results we have obtained previously provide us with an estimated upper bound of the performance of the model. Thus, we need to further analyze how the proposed model would perform when forecast temperature data is used, and whether the ensemble model is more robust to noise in forecast weather. We follow the way of modifying temperature values introduced in \cite{reis2005feature}, and consider three cases of temperature modification:

\begin{itemize}
\item{Case 1:} add Gaussian noise with mean 0 $^\text{o}$F, and standard deviation 1 $^\text{o}$F to the original temperature values before normalization.

\item{Case 2:} add Gaussian noise with mean 0 $^\text{o}$F, and change the standard deviation of case 1 to 2 $^\text{o}$F.

\item{Case 3:} add Gaussian noise with mean 0 $^\text{o}$F, and change the standard deviation of case 1 to 3 $^\text{o}$F.
\end{itemize}
For all three cases, we repeat the trials 5 times and calculate the means and standard deviations of increased MAPE compared with the case where actual temperature data is used. 

The results of increased test MAPEs for the year 2006 with modified temperature values are shown in Fig. \ref{temp}. We compare the performance of the proposed ResNetPlus model (which is an ensemble of 15 single snapshot models) with a single snapshot model trained with 700 epochs. As can be seen in the figure, the ensemble strategy greatly reduces the increase of MAPE, especially for case 1, where the increase of MAPE is 0.0168$\%$. As the reported smallest increase of MAPE for case 1 in \cite{ceperic2013strategy} is 0.04$\%$, it is reasonable to conclude that the proposed model is robust against the uncertainty of temperature for case 1 (as we use a different dataset here, the results are not directly comparable). Is is also observed that the ensemble strategy is able to reduce the standard deviation of multiple trials. This also indicates the higher generalization capability of the proposed model with the ensemble strategy.

\subsection{Probabilistic Forecasting for the Ensemble Model}

\begin{table}[!t]
\renewcommand\arraystretch{1}
\centering  % 表居中
\captionsetup{justification=centering}
\caption{Empirical Coverages of the Proposed Model with MC Dropout} 
\begin{tabular}{p{1.5cm} p{2.5cm} p{2.5cm}}
\toprule[1.5pt]
\emph{z}-score & Expected Coverage &  Empirical Coverage \\
\midrule[0.75pt]

1.000 & $68.27\%$ & $71.14\%$           \\
1.280 & $\approx80.00\%$ & $81.72\%$           \\
1.645 & $\approx90.00\%$ & $90.02\%$           \\
1.960 & $\approx95.00\%$ & $94.15\%$           \\
\bottomrule[1.5pt]
\end{tabular}
\end{table}

\begin{figure*}[!htbp]
\setlength{\abovecaptionskip}{0pt}
\centering
\includegraphics[width=17cm]{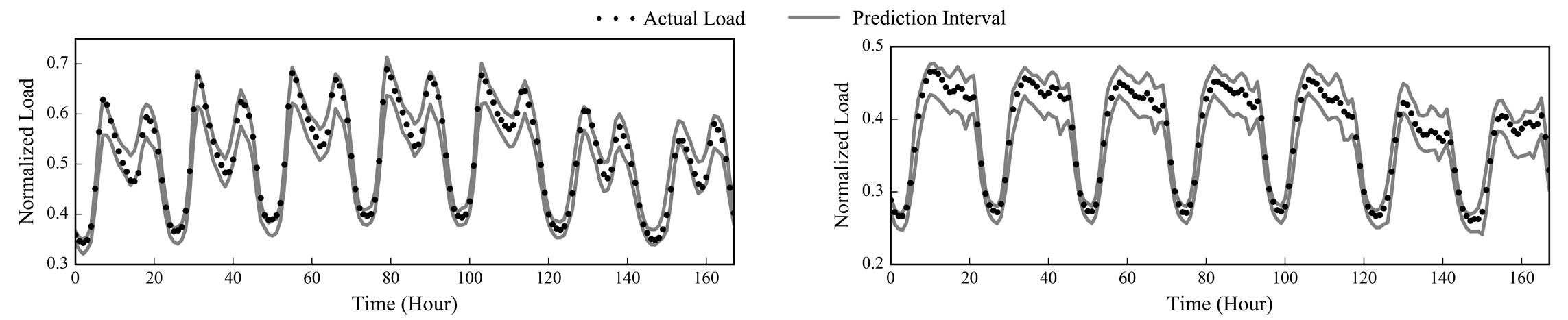}
\caption{Actual load and 95$\%$ prediction intervals for a winter week (left) and a summer week (right) of 1992 for the North-American Utility dataset. The two weeks start with February 3rd, 1992, and July 6th, 1992, respectively.}
\label{probabilistic}
\end{figure*}

We first use the North-American Utility dataset to demonstrate the probabilistic STLF by MC dropout. The last year of the dataset is used as the test set and the previous year is used for validation. Dropout with $p=0.1$ is added to the previously implemented ensemble model\footnote{the model implemented here uses ResNet instead of ResNetPlus, and the information of season, weekday/weekend distinction, and holiday/non-holiday distinction is not used. In addition, the activation function used for the residual blocks is ReLU.} except for the input layer and the output layer (dropout with $p$ ranging from 0.05 and 0.2 produce similar results, similar to the results reported in \cite{gal2016dropout}). The first term in (8) and is estimated by a single model trained with 500 epochs (with $M=100$ for (8) and $p=0.1$), and the estimated value of $\beta$ is 0.79. 

The empirical coverages produced by the proposed model with respect to different \emph{z}-scores are listed in Table V, and an illustration of the 95$\%$ prediction intervals for two weeks in 1992 is provided in Fig. \ref{probabilistic}. The results show that the proposed model with MC dropout is able to give satisfactory empirical coverages for different intervals. 

\begin{table}[!tb]
\renewcommand\arraystretch{1}
\centering  % 表居中
\captionsetup{justification=centering}
\caption{Comparison of Probabilistic Forecasting Performance Measures for the Year 2011 in the GEFCom2014 Dataset} 
\begin{tabular}{p{1.9cm} p{1.4cm} p{1.4cm} p{1.4cm}}
\toprule[1.5pt]

Model & Pinball & Winkler (50\%) & Winkler (90\%) \\
 
\midrule[0.75pt]
Lasso \cite{Ziel2016Lasso} &  $7.44$  &  -  & - \\
Ind \cite{liu2017probabilistic} & $3.22$ &  $26.35$     &   $56.38$ \\
QRA \cite{liu2017probabilistic} & $2.85$ &  $25.04$     &   $55.85$   \\
Proposed model  & $\bf{2.52}$ &  $\bf{22.41}$    &  $\bf{42.63}$ \\

\bottomrule[1.5pt]
\end{tabular}
\label{GEFCom2014}
\end{table}

In order to quantify the performance of the probabilistic STLF by MC dropout, we adopt the pinball loss and Winkler score mentioned in \cite{liu2017probabilistic} and use them to assess the proposed method in terms of coverage rate and interval width. Specifically, the pinball loss is averaged over all quantiles and hours in the prediction range, and the Winkler scores are averaged over all the hours of the year in the test set. We implement the ResNetPlus model\footnote{Five individual models are trained with a dropout rate of 0.1 and 6 snapshots are taken from 100 epochs to 350 epochs. $M$ is set to 100 for MC dropout and the first term in (7) is estimated by a single model trained with 100 epochs. The estimated value of $\beta$ is 0.77.} on the GEFCom2014 dataset and compare the results with those reported in \cite{liu2017probabilistic, Ziel2016Lasso}. Following the setting in \cite{liu2017probabilistic}, the load and temperature data from 2006 to 2009 is used to train the proposed model, the data of the year 2010 is used for validation, and the test results are obtained using data of the year 2011. The temperature values used for the input of the model are calculated as the mean of the temperature values of all 25 weather stations in the dataset. 

In Table \ref{GEFCom2014}, we present the values of pinball loss and Winkler scores for the proposed model and the models in \cite{liu2017probabilistic, Ziel2016Lasso} for the year of 2011 in the GEFCom2014 dataset. The Lasso method in \cite{Ziel2016Lasso} serves as a benchmark for methods that build regression models on the input data, and the quantile regression averaging (QRA) method in \cite{liu2017probabilistic} builds quantile regression models on sister point forecasts (the row of Ind stands for the performance of a single model). It can be seen in Table \ref{GEFCom2014} that the proposed ResNetPlus model is able to provide improved probabilistic forecasting results compared with existing methods in terms of the pinball loss and two Winkler scores. As we obtain the probabilistic forecasting results by sampling the trained neural networks with MC dropout, we can conclude that the proposed model is good at capturing the uncertainty of the task of STLF.

\section{Conclusion and Future Work}

We have proposed an STLF model based on deep residual networks in this paper. The low-level neural network with the basic structure, the ResNetPlus structure, and the two-stage ensemble strategy enable the proposed model to have high accuracy as well as satisfactory generalization capability. Two widely acknowledged public datasets are used to verify the effectiveness of the proposed model with various test cases. Comparisons with existing models have shown that the proposed model is superior in both forecasting accuracy and robustness to temperature variation. We have also shown that the proposed model can be directly used for probabilistic forecasting when MC dropout is adopted.

A number of paths for further work are attractive. As we have only scratched the surface of state-of-the-art of deep neural networks, we may apply more building blocks of deep neural networks (e.g., CNN or LSTM) into the model to enhance its performance. In addition, we will further investigate the implementation of deep neural works for probabilistic STLF and make further comparisons with existing methods.

{
\small
\bibliographystyle{IEEEtran}%
\bibliography{ISO.bib}
}

\begin{IEEEbiography} [{\includegraphics[width=1in,height=1.25in,clip,keepaspectratio]{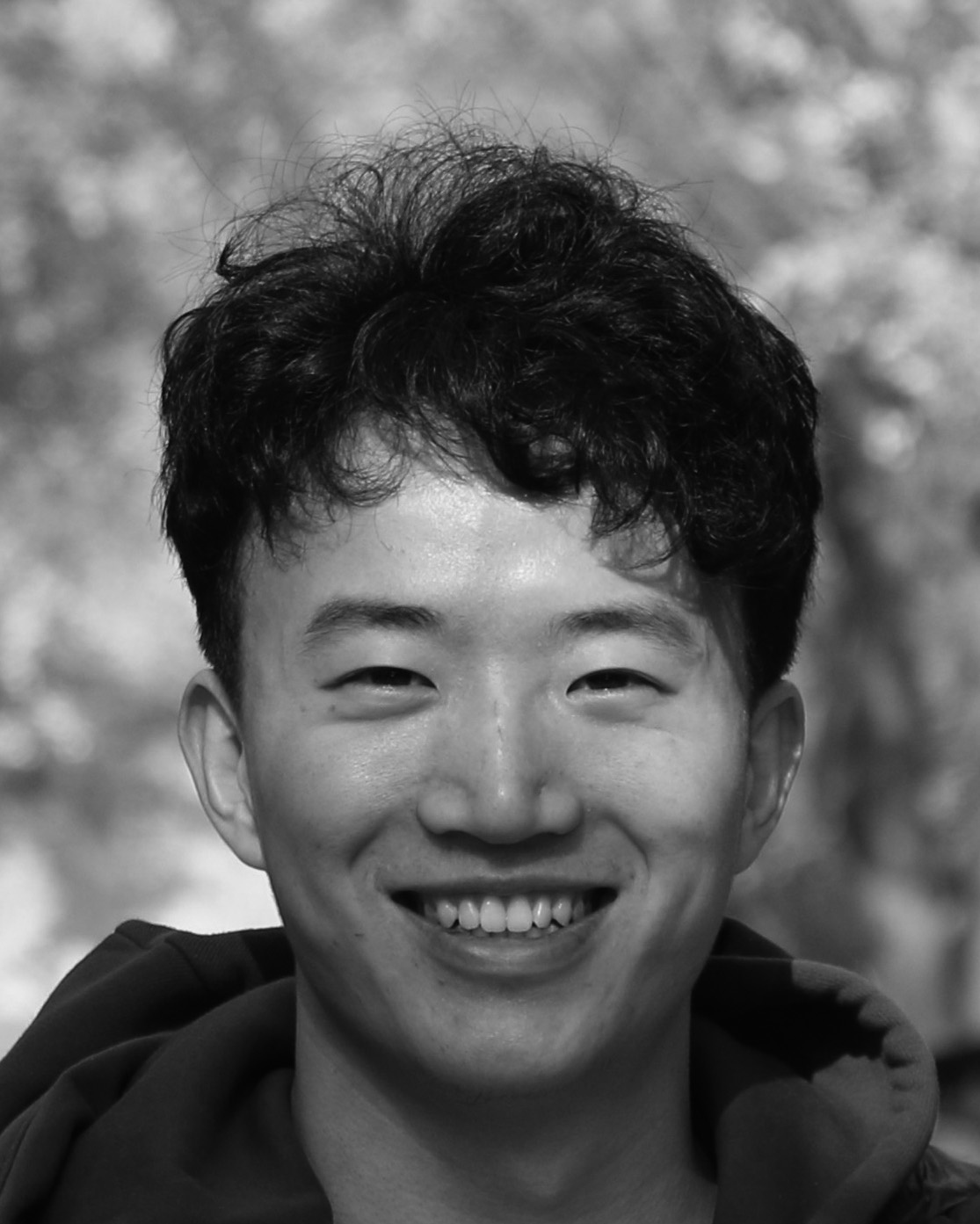}}]
{Kunjin Chen}received the B.Sc. degree in electrical engineering from Tsinghua University, Beijing, China, in 2015. Currently, he is a Ph.D. student with the Department of Electrical Engineering. 

His research interests include applications of machine learning and data science in power systems.
\end{IEEEbiography}

\vspace{-10 mm}

\begin{IEEEbiography} [{\includegraphics[width=1in,height=1.25in,clip,keepaspectratio]{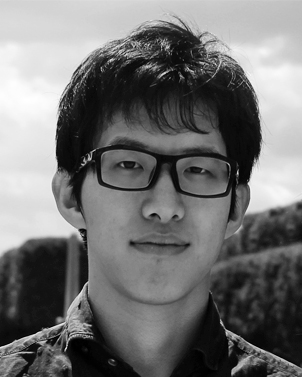}}]
{Kunlong Chen} received the B.Sc. degree in electrical engineering from Beijing Jiaotong University, Beijing, China and the engineering degree from CentraleSup\'{e}lec, Paris, France. He is currently pursuing the M.Sc. degree with the Department of Electrical Engineering in Beijing Jiaotong University. 

His research interests include applications of statistical learning techniques in the field of electrical engineering.
\end{IEEEbiography}

\vspace{-10 mm}

\begin{IEEEbiography} [{\includegraphics[width=1in,height=1.25in,clip,keepaspectratio]{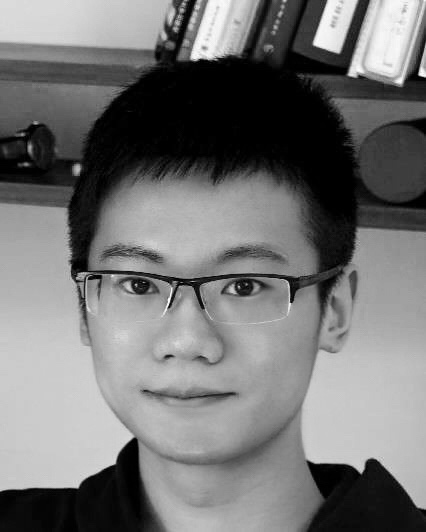}}]
{Qin Wang} received his B.Eng degree in electrical engineering from Tsinghua University in 2015. Currently, he is a Master student at ETH Zurich. 

His reserach interests include computer vision and deep learning applications.
\end{IEEEbiography}

\vspace{-10 mm}

\begin{IEEEbiography} [{\includegraphics[width=1in,height=1.25in,clip,keepaspectratio]{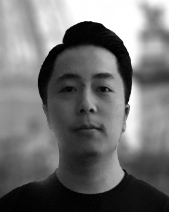}}]
{Ziyu He} recieved his B.E. degree from Zhejiang University in 2015 and his M. S. degree from Columbia University in 2017. He is currently a Ph.D. student in the department of industrial and systems engineering at University of Southern California. 

His research interests are optimization, machine learning and their applications in energy.
\end{IEEEbiography}

\vspace{-10 mm}

\begin{IEEEbiography} [{\includegraphics[width=1in,height=1.25in,clip,keepaspectratio]{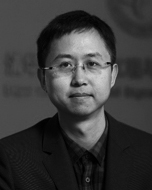}}]
{Jun Hu} (M'10) received his B.Sc., M.Sc., and Ph.D. degrees in electrical engineering from the Department of Electrical Engineering, Tsinghua University in Beijing, China, in July 1998, July 2000, July 2008. 

Currently, he is an associate professor in the same department. His research fields include overvoltage analysis in power system, sensors and big data, dielectric materials and surge arrester technology.
\end{IEEEbiography}

\vspace{-10 mm}

\begin{IEEEbiography} [{\includegraphics[width=1in,height=1.25in,clip,keepaspectratio]{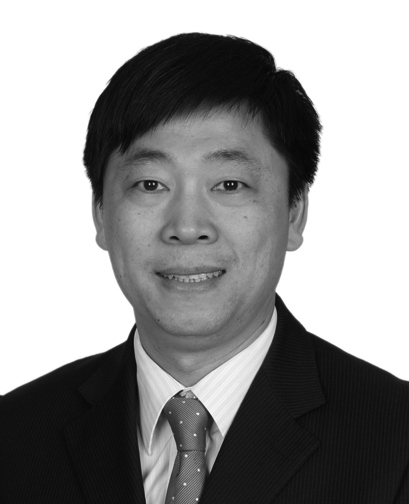}}]
{Jinliang He} (M'02--SM'02--F'08) received the B.Sc. degree from Wuhan University of Hydraulic and Electrical Engineering, Wuhan, China, the M.Sc. degree from Chongqing University, Chongqing, China, and the Ph.D. degree from Tsinghua University, Beijing, China, all in electrical engineering, in 1988, 1991 and 1994, respectively.

He became a Lecturer in 1994, and an Associate Professor in 1996, with the Department of Electrical Engineering, Tsinghua University. From 1997 to 1998, he was a Visiting Scientist with Korea Electrotechnology Research Institute, Changwon, South Korea, involved in research on metal oxide varistors and high voltage polymeric metal oxide surge arresters. From 2014 to 2015, he was a Visiting Professor with the Department of Electrical Engineering, Stanford University, Palo Alto, CA, USA. In 2001, he was promoted to a Professor with Tsinghua University. He is currently the Chair with High Voltage Research Institute, Tsinghua University. He has authored five books and 400 technical papers. His research interests include overvoltages and EMC in power systems and electronic systems, lightning protection, grounding technology, power apparatus, and dielectric material.
\end{IEEEbiography}

\end{document}